\documentclass[runningheads]{llncs}
\usepackage{graphicx}
\begin{document}
\title{Multi-label Thoracic Disease Image Classification with Cross-Attention Networks}
%
%\titlerunning{Abbreviated paper title}
% If the paper title is too long for the running head, you can set
% an abbreviated paper title here
%
\author{Congbo Ma\inst{1, 2} \and % index{Ma, Congbo}
Hu Wang\inst{1, 3} \and % index{Wang, Hu}
Steven C.H. Hoi\inst{1, 4}} % index{Hoi, Steven C.H.}
\authorrunning{C. Ma et al.}
% First names are abbreviated in the running head.
% If there are more than two authors, 'et al.' is used.
%
\institute{Singapore Management University, Singapore
 \and South China University of Technology, Guangzhou, China
 \and The University of Adelaide, Adelaide, Austrilia
 \and Salesforce Research Asia, Singapore
}

\maketitle              % typeset the header of the contribution
\begin{abstract}

Automated disease classification of radiology images has been emerging as a promising technique to support clinical diagnosis and treatment planning. Unlike generic image classification tasks, a real-world radiology image classification task is significantly more challenging as it is far more expensive to collect the training data where the labeled data is in nature multi-label; and more seriously samples from easy classes often dominate; training data is highly class-imbalanced problem exists in practice as well. To overcome these challenges, in this paper, we propose a novel scheme of Cross-Attention Networks (CAN) for automated thoracic disease classification from chest x-ray images, which can effectively excavate more meaningful representation from data to boost the performance through cross-attention by only image-level annotations. We also design a new loss function that beyond cross entropy loss to help cross-attention process and is able to overcome the imbalance between classes and easy-dominated samples within each class. The proposed method achieves state-of-the-art results.

\end{abstract}

\section{Introduction}

Chest diseases are constantly a big threat to people's health. Early diagnosis and treatment of chest diseases are very important. Computer aided x-ray analysis is an effective way to diagnose chest diseases. Wang et al.~\cite{Chestx-ray14} constructed one of the largest public Chest X-Ray14 datasets. After that, Rajpurkar et al.~\cite{chexnet} proposed a 121 layers convolutional neural network trained on Chest X-Ray14 dataset. Later on, Li et al .~\cite{Lifeifei} proposed a unified structure that can perform disease identification and localization simultaneously. They used both normal labelled x-ray images and those with disease location annotations. However, annotated data are quite expensive to acquire and heavily dependent on expert experience. Recently, a large-scale chest x-ray dataset CheXpert~\cite{chexpert} came out. For chest x-ray image processing, data is usually multi-labeled and for each label, positive-negative samples are often imbalanced and easy samples are usually in a dominant position, which usually result in poor performance. The general training data number upsampling or downsampling approach and cross entropy loss may not be an ideal solution for multi-label imbalance classification problem.

To this end, we propose Cross-Attention Networks with a new loss function to tackle imbalanced multi-label x-ray chest diseases classification from two different angles: (1) From image processing angle, we design a flexible end-to-end training Cross-Attention Network architecture which can effectively excavate more meaningful representation with only image-level annotations. By using hadamard product of two feature maps, the proposed structure could effectively eliminate attention noises. (2) From a learning and optimization perspective, we proposed a new loss function which consists of an attention loss that could facilitate the model to learn better representations and a multi-label balance loss to reduce imbalance between positive and negative classes within each disease and dominated easy samples. We have also used image-level supervised localization to validate our model which is able to localize at high risk pathogenic areas in a better manner. We make thorough experiments on Chest X-Ray14 and CheXpert datasets to evaluate the effectiveness of our different proposed components. Our method achieves the state-of-the-art results.

\section{Model}

\subsection{Cross-Attention Networks}

\textbf{Feature extraction networks.} After raw data pre-processing, the images are pumped into two feature extraction networks and gone through convolutional layers which are worked as feature extractors. Our proposed cross-attention method is flexible and the two networks can be easily substituted by other backbone networks. Different from standard CNNs, instead of pushing extracted features into activation layers and global average pooling layers, we keep these abstract image-level features for later use.

\begin{figure}
\centering
\hspace*{-1.0cm}
\includegraphics[width=1.15\textwidth]{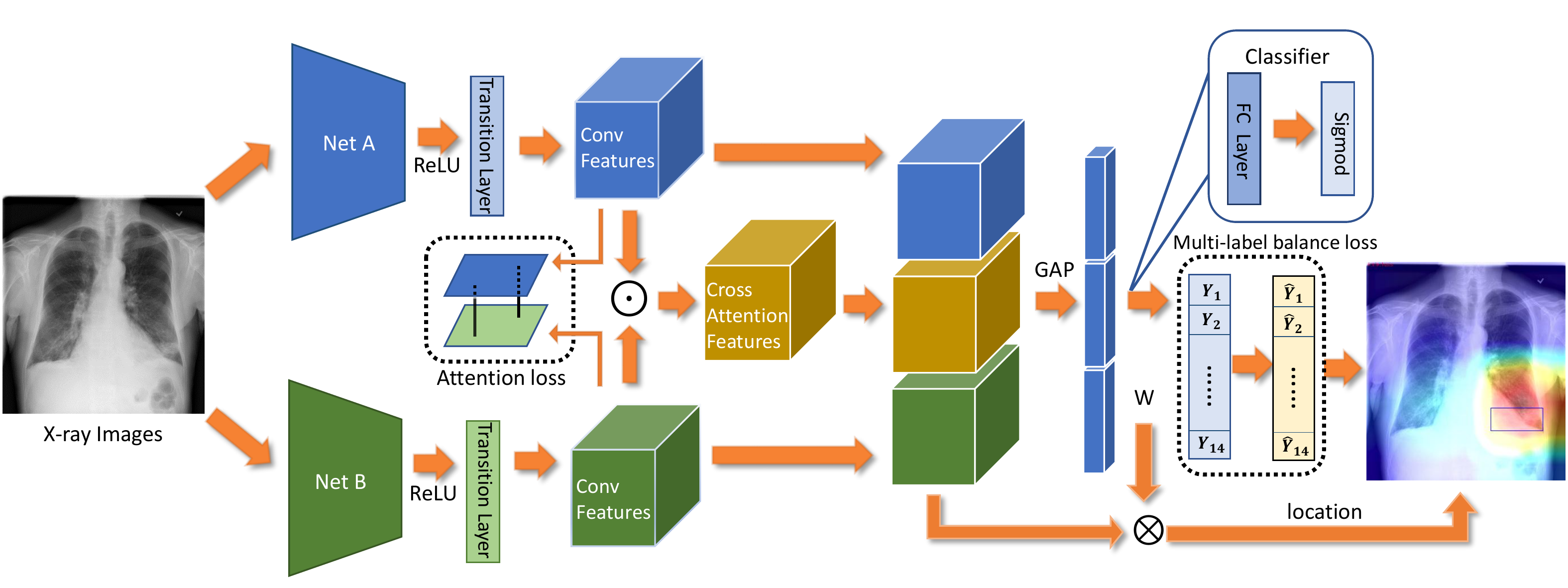}
\caption{The proposed Cross-Attention Networks. (1) Data are inputted through ReLU layers to eliminate the interference of negative activated values; (2) Transition layers play the role to transform two groups of feature maps into the same dimension; (3) The element-wise hadamard product is used to get cross-attention feature maps which urges networks to only focus on the area that have high pathogenic probabilities; (4) Dashed boxes represent attention loss and multi-label balance loss.}
\end{figure}

\textbf{Cross-attention model.} We proposed a new cross-attention model that two networks have attention on each other, which could generate more meaningful representations. In Figure 1, after we get the features by pumping images into two networks which have different initialization or structures, the two feature maps will go through two ReLU layers respectively to ensure that negative activation values will not interfere the cross-attention features. Then these feature maps will be inputted to a transition layer to transform two groups of features into the same shape. Later on, instead of outer-producting two large tensors as self-attention ~\cite{self-attention} methods by building attention for each pair of data point, we used element-wise hadamard product to get cross-attention feature maps. Cross-attention structure enables networks to get attention on each other and eliminate noises, which means the Cross-Attention Networks have the ability to only focus on the areas that have high pathogenic chances. Then the cross-attention feature maps will be concatenated with two output tensors.

\begin{center}
$Feature_{all\_cat} = Concat(F_{CA},F_{A},F_{B}), where F_{CA} = F_{A}\odot F_{B}$
\end{center}

where $F_{A}$,  $F_{B}$ represents the feature maps of Network A and Network B, $F_{CA}$ represents cross-attention feature maps. $Concat()$ function represents concatenate function. By using hadamard product, the cross-attention feature maps $F_{CA}$ would be activated on those areas that only if the two groups of feature maps $F_{A}$ and $F_{B}$ are both activated, which eliminate the randomness of outliers. When the loss is back propagated, two networks affect each others' gradient flows which also enable two networks update in a more collaborative way.

\textbf{Transition layers and classifier.} The transition layers are to transform two groups of feature maps into the same dimension. We use $tran_N = Min(C_{A},C_{B})$ number of $1*1$ convolutional kernels to transform these two group of feature maps into the same shape. where $C_{A}$ and $C_{B}$ denote the channel numbers of different feature maps. By using transition layers, we ensure the dimension of two feature maps are the same and also add further non-linearity to our cross-attention model.

In Figure 1, after getting the all concatenated cross-attention feature maps, we input these feature maps into a global average pooling layer and then the classifier. Our model is designed to have different binary classification for different labels. For each label, the disease probability score is calculated through a sigmoid function. So our classifier consists of a fully connected layer and a sigmoid layer to predict the probability of images which may have thoracic diseases.

\subsection{Loss design}

\textbf{Attention loss.} In order to further excavate more meaningful features in proposed cross-attention model, we define a new attention loss. In figure 1, we first fetch all feature maps from the last block before the global average pooling of two networks. Each pixel in these feature maps encodes different spatial information of the original image. Then we channel-wise sum up these features maps to form a pathogenic attention map. Different feature maps within one network have different activation areas, however, majority of these features would concentrate on the "right" areas. By getting the summation of these feature maps within one network, we would coarsely get a pathogenic attention map.

By feeding through a resize layer, we get two same size $1*H'*W'$ pathogenic attention maps from two networks, where $H'$ and $W'$ represents the height and width of each feature map. After that, we normalized these two pathogenic attention maps by dividing the max value of each attention map to ensure each pixel has the same range of values. Then we calculate the L2 loss of the two pathogenic attention maps. Through attention loss, we set a constraint to urge two networks to find mutual pathogenic areas which would smooth the cross-attention process. The formulation of attention loss is given as below:

\begin{center}
$L_{att} = || norm(\sum_{n=1}^{N} f_{n}) - norm(\sum_{m=1}^{M} f_{m}) ||_2$
\end{center}

where $N$ and $M$ denote the total size of feature channels within one network. $n$ and $m$ represent the index of each feature map. $f_{n}$ and $f_{m}$ are different feature maps from two networks respectively.

\textbf{Multi-label balance loss.} Poor performance can be easily caused by imbalance between classes, which is not ideal for standard cross entropy loss to solve. And easy-dominated samples hamper models to learning genuine discriminated features. Inspired by Lin et al.~\cite{focalloss}, we designed a loss that extends the focal loss to multi-label setting with balance factors. To our best of knowledge, it is the first work to do that. Through this loss, our model can not only handle the imbalance between positive-negative samples within each disease, but also excavate more meaningful representation out of dominated easy samples. We sum up the balance loss from each disease to represent multi-label balance loss.

\begin{center}
$L_{bal} = \sum_{l=1}^{L} -w_{l-}\cdot [1-p_{l}(Y=0|X)]^{\gamma }(1-y)logp_{l}(Y=0|X) +$
$-w_{l+}\cdot [1-p_{l}(Y=1|X)]^{\gamma }ylogp_{l}(Y=1|X)$
\end{center}

where $L$ is the total number of multiple labels and $l$ means each label of different type of diseases. $w_{l+}=\frac{\left |N_{l}  \right |}{\left | P_{l} \right | + \left | N_{l} \right |}$ and $w_{l-}=\frac{\left |P_{l}  \right |}{\left | P_{l} \right | + \left | N_{l} \right |}$, which means balance factor to eliminate positive-negative sample imbalance between each binary classes. $P_{l}$ and $N_{l}$ represent numbers of positive and negative samples of a certain disease. Parameter $\gamma$ controls the curvature and shape of the multi-label balance loss, which may have better performance to mine "hard" examples to contribute more to the model training in some $\gamma$ settings, especially when "easy" examples dominate the dataset greatly.

\textbf{Cross-attention loss function.} The cross-attention loss function is a combined loss function which is:

\begin{center}
$L = \alpha L_{att} + L_{bal}$. 
\end{center}

where $\alpha$ represents the trade-off factor between attention loss and multi-label balance loss. The attention loss could force the model to focus on pathogenic areas more accurately. Multi-label balance loss helps the model eliminate positive-negative sample imbalance problem and easy sample dominated problem within each disease.

\section{Experiments}
\subsection{Experiments on Chest X-Ray14 Dataset}
\subsubsection{Experimental Settings}

We validate our proposed Cross-Attention Networks on Chest X-Ray14 dataset~\cite{Chestx-ray14} and follow the official train-val and test data split to keep fair comparison. We further split the official train-val set into 78485 images for training and 8039 images for validation and ensure no patient overlap among these three sets. We downscale input images to the size of 256*256 and random crop to 224*224 with the batch-size of 96 and initial learning rate of 0.001. We set the dropout rate of last fully connected layer to 0.2 and loss trade-off factor to 0.01. In the multi-label balance loss, we empirically set $\gamma$ to 2. All experiments are evaluated in terms of AUROC values.

\subsubsection{Model comparison}

The quantitative performance of models comparison is demonstrated in Table 1. CAN$_1$ represents the Cross-Attention Networks $model_1$ with densenet121 and densenet169 as its backbone networks and CAN$_2$ means the Cross-Attention Networks $model_2$ with two densenet121. The two networks within our model are initialized with different weights which are warmed up at the dataset. Since the experiments of Li et al.~\cite{Lifeifei} is not done on the official split~\cite{Chestx-ray14} and they leverage extra location labels to train, in order to have a fair comparison, their results will not participate in best result comparison in each row (marked in bold font). 

\begin{table}
\centering
\caption{Results comparison between different methods on Chest X-Ray14 Dataset}
\scalebox{0.8}{
\begin{tabular}{p{2.2cm} | p{1.5cm} p{1.5cm} p{1.5cm} p{1.5cm} p{1.5cm} p{1.5cm} | p{1.3cm} p{1.1cm}}
\hline
Diseases      & Wang etc.~\cite{Chestx-ray14}    & Yao etc.~\cite{yao-paper}     & CheXNet ~\cite{chexnet} & Li etc.~\cite{Lifeifei}* & Guendel etc.~\cite{Guendel-paper} & Xia etc.~\cite{xiayong-paper} & CAN$_1$  & CAN$_2$        \\ \hline
Split by Wang & Yes            & Yes            & Yes     & No          & Yes            & Yes        & Yes     & Yes            \\
Image resize    & 256*256        & 256*256        & 256*256 & 512*512     & 256*256        & 256*256    & 256*256 & 256*256        \\\hline
Atelectasis   & 0.773          & 0.733          & 0.759   & 0.800       & 0.767          & 0.743      & 0.772   & \textbf{0.777} \\
Cardiomegaly  & 0.854          & 0.856          & 0.871   & 0.870       & 0.883          & 0.875      & 0.894   & \textbf{0.894} \\
Effusion      & \textbf{0.861} & 0.806          & 0.821   & 0.870       & 0.828          & 0.811      & 0.828   & 0.829          \\
Infiltration  & 0.636          & 0.673          & 0.700    & 0.700       & \textbf{0.709} & 0.677      & 0.703   & 0.696          \\
Mass          & 0.761          & 0.718          & 0.810   & 0.830       & 0.821          & 0.783      & 0.830   & \textbf{0.838} \\
Nodule        & 0.664          & \textbf{0.777} & 0.759   & 0.750       & 0.758          & 0.698      & 0.762   & 0.771          \\
Pneumonia     & 0.664          & 0.684          & 0.718   & 0.670       & \textbf{0.731} & 0.696      & 0.721   & 0.722          \\
Pneumothorax  & 0.799          & 0.805          & 0.848   & 0.870       & 0.846          & 0.810      & 0.856   & \textbf{0.862} \\
Consolidation & \textbf{0.770} & 0.711          & 0.741   & 0.800       & 0.745          & 0.726      & 0.756   & 0.750          \\
Edema         & \textbf{0.861} & 0.806          & 0.844   & 0.880       & 0.835          & 0.833      & 0.846   & 0.846          \\
Emphysema     & 0.736          & 0.842          & 0.891   & 0.910       & 0.895          & 0.822      & 0.892   & \textbf{0.908} \\
Fibrosis      & 0.739          & 0.743          & 0.810   & 0.780       & 0.818          & 0.804      & 0.824   & \textbf{0.827} \\
PT            & 0.749          & 0.724          & 0.768   & 0.790       & 0.761          & 0.751      & 0.773   & \textbf{0.779} \\
Hernia        & 0.746          & 0.775          & 0.867   & 0.770       & 0.896          & 0.900      & 0.932   & \textbf{0.934} \\\hline
Average       & 0.758          & 0.761          & 0.801   & 0.806       & 0.807          & 0.781      & 0.814   & \textbf{0.817} \\ \hline
\end{tabular}}
\end{table}

From Table 2, the cross-attention model achieved the best result in terms of average AUROC scores for most individual disease cases. The improvements are remarkable especially for those diseases with extremely scarce positive samples. For example, ``Hernia" only has 227 images (\%0.202) and ``Fibrosis" only contains 1686 images (\%1.504) in the dataset, the proposed CAN$_2$ model obtained 0.932 and 0.827 in terms of the AUROC score respectively, which is much better than its competitors. This is because our cross-attention model with the newly designed cross-attention loss function penalizes those hard examples and can have better differentiate between positive/negative in a good manner.

\subsubsection{Ablation study}

To evaluate the effectiveness of our different proposed components, we also implement them separately in ablation study to monitor how they influence the performance. In Table 3, "121" and "169" represent Densenet121 and Densenet169. "had" means element-wise hadamard product cross-attention. $L_{bce}$, "$L_{bal}$" and "$L_{att}$" are binary cross entropy loss, multi-label balance loss and attention loss. "all\_cat" is the operation to get all concatenated cross-attention feature maps. In order to have a fair comparison, we have not only done the experiments to compare with existing methods, but also compared with different feature aggregation methods with exact parameter numbers.  "add" and "max" means element-wise addition or maximum operation between two feature maps respectively, and these two methods have the same parameters with our proposed cross-attention networks.

\begin{table}
\centering
\caption{Results comparison between different methods}
\scalebox{0.8}{
\begin{tabular}{p{4cm}l|ll}
\hline
Methods                         &       & Methods                                   &                \\ \hline
121+$L_{bce}$                   & 0.801 & 121+169(add+all\_cat)+$L_{bal}$+$L_{att}$ & 0.810          \\
121+$L_{bal}$                   & 0.806 & 121+169(max+all\_cat)+$L_{bal}$+$L_{att}$ & 0.810          \\
169+$L_{bal}$                   & 0.806 & 121+169(had)+$L_{bal}$+$L_{att}$          & 0.811          \\
121+169(had)+$L_{bal}$          & 0.809 & Cross-attention $model_1$(CAN$_1$)            & 0.814          \\
121+169(had+all\_cat)+$L_{bal}$ & 0.810 & Cross-attention $model_2$(CAN$_2$)            & \textbf{0.817} \\ \hline
\end{tabular}}
\end{table}

From Table 2, we found that our CAN model boosts the performance in such imbalanced dataset. By using "hadamard" produce of two feature maps, the average AUROC score increases from 0.806 to 0.809. Because the randomness of “poor” activation and outliers are eliminated. Also, the model can be updated through back propagating gradients of each other in a more collaborative way; however, "add" and "max" operations do not have this attribute, so even with all concatenate feature maps and cross-attention loss function, the results are both stagnated at 0.810. With the help of proposed attention loss, the AUROC score is lifted to 0.811. Because attention loss drags the attention of two networks closer which would facilitate cross-attention process. Our CAN model achieves AUROC of 0.814 and 0.817, the new state-of-the-art results.

\subsubsection{Image-level supervised disease localization}

To better understand our model, disease localization heat maps are generated by only image-level supervised labels~\cite{cam}. Since we used Global Average Pooling as our last pooling layer, we directly sum up the multiplication of weights and feature maps between pooling layer and fully connected layer to localize.

\begin{figure}
\begin{center}
\includegraphics[width=0.9\textwidth]{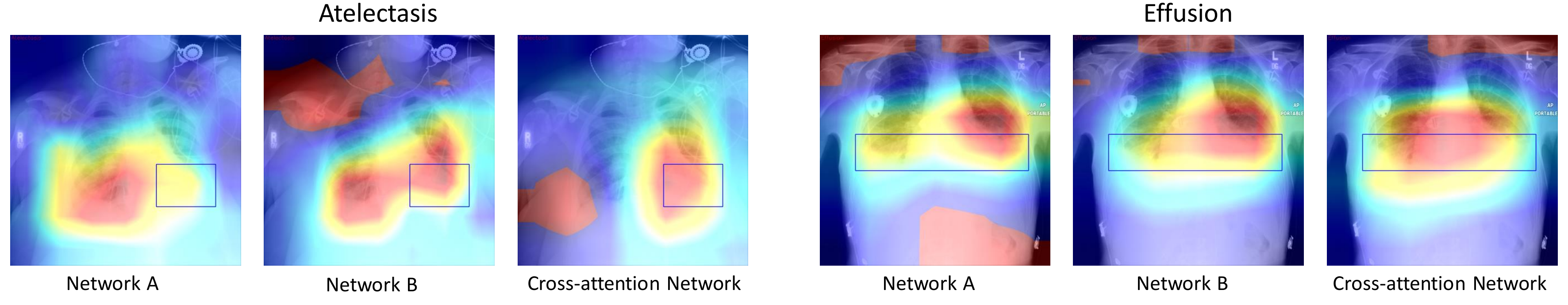}
\caption{The results of the localization of chest diseases.}
\end{center}
\end{figure}

In Figure 2, the heat maps localize on high-probability disease areas and the blue bounding boxes indicate the ground-truth location of the diseases. Our proposed Cross-Attention Networks provide better performance to focus on high-probability disease areas in a better manner than using the networks separately. This means that Cross-Attention Networks enable model updates in a collaborative way and could have better feature representations.

\subsection{Experiments on CheXpert Dataset}
\subsubsection{Experimental Settings}

We also validate our proposed Cross-Attention Networks on newly released dataset CheXpert~\cite{chexpert} with "Frontal views" and "Lateral views", and keep the exact same settings of compared methods and our method throughout all experiments. In the CheXpert experiments, in order to have enough data for testing, we split 222914 images for training and the other 734 images for testing, and we further ensure no patient overlap among them.

\subsubsection{Model comparison}

In our experiments, we keep the consistent experimental settings by only using U(uncertain)-Ones introduced by the CheXpert paper~\cite{chexpert} --- mapping all instances of the uncertain label to 1, and we test all models on 14 labels by either just using "Frontal views" or treating "Frontal views" and "Lateral views" equally, which means we do not select the best uncertainty approach for each disease and do not use maximum probability of the observations across the views or other bells and whistles. We further compare the results obtained by "Add" and "Max" operations. Table 3 shows the experimental results comparison on 14 labels classification tasks.

\begin{table}
\centering
\caption{Results comparison on 14 labels classification tasks on CheXpert dataset}
\scalebox{0.8}{
\begin{tabular}{l|llll|llll}
\hline
Experiments      & \multicolumn{4}{l|}{Frontal Views Only}                    & \multicolumn{4}{l}{Frontal + Lateral (Equally)}                            \\ \hline
Labels           & CheXNet & Add            & Max            & CAN      & CheXNet        & Add            & Max            & CAN      \\ \hline
Atelectasis      & 0.659   & \textbf{0.683} & 0.697          & 0.667          & 0.707          & 0.678          & 0.691          & \textbf{0.713} \\
Cardiomegaly     & 0.775   & \textbf{0.781}          & 0.760          & 0.773          & 0.775          & 0.777          & 0.757          & \textbf{0.790} \\
Consolidation    & 0.702   & \textbf{0.735} & 0.729          & 0.732          & 0.755          & 0.746          & 0.730          & \textbf{0.757} \\
Edema            & 0.827   & \textbf{0.846} & 0.843          & 0.840          & 0.863          & 0.858          & \textbf{0.867} & 0.861          \\
Enlarged Cardio  & 0.551   & 0.505          & 0.518          & \textbf{0.552} & 0.531          & 0.491          & \textbf{0.568} & 0.555          \\
Fracture         & 0.616   & \textbf{0.732} & 0.710          & 0.722          & 0.588          & 0.638          & 0.608          & \textbf{0.735} \\
Lung Lesion      & 0.704   & 0.775          & \textbf{0.831} & 0.757          & 0.710          & 0.778          & 0.741          & \textbf{0.805} \\
Lung Opacity     & 0.767   & 0.741          & 0.768          & \textbf{0.788} & \textbf{0.784} & 0.764          & 0.770          & 0.783          \\
No Finding       & 0.887   & 0.879          & 0.883          & \textbf{0.893} & \textbf{0.872} & 0.865          & 0.865          & 0.859          \\
Pleural Effusion & 0.860   & 0.887          & 0.891          & \textbf{0.892} & 0.874          & 0.887          & 0.881          & \textbf{0.892} \\
Pleural Other    & 0.607   & 0.699          & 0.647          & \textbf{0.711} & 0.710          & \textbf{0.718} & 0.710          & 0.680          \\
Pneumonia        & 0.641   & 0.674          & 0.671          & \textbf{0.710} & 0.535          & 0.601          & 0.647          & \textbf{0.666} \\
Pneumothorax     & 0.807   & 0.812          & 0.823          & \textbf{0.824} & \textbf{0.842} & 0.826          & 0.809          & 0.836          \\
Support Devices  & 0.869   & 0.882          & 0.877          & \textbf{0.889} & 0.899          & 0.879          & 0.879          & \textbf{0.913} \\ \hline
Average          & 0.734   & 0.759          & 0.760          & \textbf{0.768} & 0.746          & 0.750          & 0.752          & \textbf{0.775} \\ \hline
\end{tabular}}
\end{table}

From Table 3, We can find that the cross-attention model achieved the best AUROC scores for most individual disease cases in whether "Frontal Views Only" or "Frontal + Lateral Views" setting.

\section{Conclusion}

This paper proposed an end-to-end trainable Cross-Attention Networks (CAN) scheme for multi-label x-ray chest diseases classification. The Cross-Attention Networks not only utilize the attention loss for better attention at the regions of interest, but also overcome the positive-negative sample imbalance and easy sample dominated problems with multi-label balanced loss. Cross-Attention Networks can effectively excavate meaningful representation from data by having attention on each other and updating models in a more collaborative way. Quantitative and qualitative results demonstrate the state-of-the-art performance of our method.

\section{Acknowledgement}

Most of this work were done when the authors worked at Singapore Management University.

%
% ---- Bibliography ----
%
% BibTeX users should specify bibliography style 'splncs04'.
% References will then be sorted and formatted in the correct style.
%
% \bibliographystyle{splncs04}
% \bibliography{mybibliography}
%

\end{document}